\documentclass[10pt,conference]{IEEEtran}

\usepackage{cite}
\usepackage{amsmath,amssymb,amsfonts}
\usepackage{graphicx}
\usepackage[caption=false,font=footnotesize]{subfig}
\usepackage{booktabs}
\usepackage{multirow}
\usepackage{array}
\usepackage{url}
\usepackage[hidelinks]{hyperref}
\usepackage{algorithm}
\usepackage{algorithmic}
\usepackage{xcolor}

\title{Optimistic Feasible Search for Closed-Loop Fair Threshold Decision-Making}

\author{
\IEEEauthorblockN{Du Wenzhang}
\IEEEauthorblockA{Dept. of Computer Engineering\\
Mahanakorn University of Technology, International College (MUTIC)\\
Bangkok, Thailand\\
dqswordman@gmail.com}
}

\begin{document}
\maketitle

% ===================== Abstract =====================
\begin{abstract}
Closed-loop decision-making systems (e.g., lending, screening, or recidivism risk assessment) often operate under fairness and service constraints while simultaneously inducing feedback effects: decisions change who appears in the future, yielding non-stationary data and potentially amplifying disparities.
We study online learning of a one-dimensional threshold policy from bandit feedback under demographic parity (DP) and (optionally) service-rate constraints.
The learner observes only a scalar score each round and selects a threshold; reward and constraint signals are revealed for the chosen threshold only.

We propose Optimistic Feasible Search (OFS), a simple grid-based method that maintains confidence bounds for reward and constraint residuals for each candidate threshold.
At each round, OFS selects the threshold that appears feasible under confidence bounds and, among those, maximizes optimistic reward; if no threshold appears feasible, OFS selects the threshold minimizing optimistic violation.
This design directly targets feasible high-utility thresholds and is particularly effective for low-dimensional, interpretable policy classes where discretization is natural.

We evaluate OFS on (i) a synthetic closed-loop benchmark with stable contraction dynamics and (ii) two semi-synthetic closed-loop benchmarks grounded in German Credit and COMPAS, constructed by training a score model and feeding group-dependent acceptance decisions back into population composition.
Across all environments, OFS achieves substantially higher reward with dramatically smaller cumulative constraint violation than unconstrained and primal--dual bandit baselines.
Moreover, OFS is near-oracle: its steady-state reward gap to the best feasible fixed threshold is below 0.003 across benchmarks.
All experiments are reproducible and organized with double-blind friendly relative outputs.
\end{abstract}


% ===================== I. Introduction =====================
\section{Introduction}
Algorithmic decision systems increasingly mediate access to opportunities and resources in high-stakes settings such as lending and criminal justice.
A core difficulty is that these systems are closed-loop: decisions affect the future population that the system will encounter (selection effects, behavioral responses, or measurement shifts), creating non-stationarity and potentially reinforcing disparities.
At the same time, regulations and organizational policies often require fairness constraints such as demographic parity (DP) \cite{dwork2012fairness,feldman2015certifying,agarwal2018reductions} or related notions \cite{hardt2016equality,kleinberg2017inherent}.

This paper focuses on a minimal but practically relevant closed-loop learning problem:
a decision-maker observes a scalar score (e.g., a model logit) and chooses a threshold $\theta$ (accept if score $\ge \theta$).
Threshold policies are widely used for their transparency and ease of auditing.
However, learning a good threshold online under constraints is challenging because the environment is non-stationary and feedback is bandit-style.

\textbf{Goal.}
We aim to maximize long-run utility while keeping constraint violations small, using only bandit feedback.
We specifically target the regime where the policy class is low-dimensional (here 1D thresholds), so a discrete search strategy with rigorous uncertainty estimates is viable.

\textbf{Contributions.}
\begin{itemize}
  \item \textbf{Method:} We propose Optimistic Feasible Search (OFS), an optimism-based feasible screening method for constrained closed-loop threshold learning, inspired by optimism in bandits \cite{auer2002ucb}. OFS is simple, interpretable, and works directly on discretized thresholds (Section~\ref{sec:ofs}).
  \item \textbf{Benchmarks:} We provide a synthetic closed-loop benchmark (MVE-S) with stable contraction dynamics and two semi-synthetic benchmarks derived from German Credit and COMPAS via score-model-based logits and feedback-induced population shifts (Section~\ref{sec:benchmarks}).
  \item \textbf{Evidence:} Using a unified experimental suite and fully reproducible relative outputs, OFS dominates strong baselines in reward and cumulative violations and is near-oracle across all benchmarks (Section~\ref{sec:experiments}, Tables~\ref{tab:main_results}, \ref{tab:oracle_gap} and Figs.~\ref{fig:mve_curves}--\ref{fig:oracle_curves}).
\end{itemize}

\textbf{Related context.}
Our work connects constrained online learning \cite{zinkevich2003online,mahdavi2012online,achiam2017cpo} and bandit/zeroth-order optimization \cite{flaxman2005online,spall1992spsa,nesterov2017random}.
It is also motivated by growing recognition of performative or feedback-driven effects in prediction and decision pipelines \cite{perdomo2020performative}.
Compared to many fairness-aware learning approaches that assume i.i.d.\ data \cite{agarwal2018reductions}, we emphasize closed-loop non-stationarity and provide benchmark constructions to study it.

% ===================== II. Problem Setup =====================
\section{Problem Setup}\label{sec:setup}
At each round $t=1,\dots,T$, the environment produces a batch of $n$ individuals (we use $n{=}256$ in experiments) with a sensitive attribute $a\in\{0,1\}$, a scalar score $s$, and an outcome label $y\in\{0,1\}$.
The decision-maker selects a threshold policy
\begin{equation}
\pi_\theta(s) = \mathbb{I}\{ s \ge \theta \},
\end{equation}
yielding decisions $d\in\{0,1\}$ (accept/reject).
A bandit feedback signal is observed: reward and constraint residuals for the chosen threshold.

\subsection{Utility}
We consider a standard acceptance utility with false-positive cost $c_{\mathrm{fp}}>0$:
\begin{equation}
r_t(\theta) = \mathbb{E}\big[d\cdot (y - c_{\mathrm{fp}}(1-y))\big],
\end{equation}
estimated by the batch mean.
This captures the trade-off between approving beneficial cases ($y{=}1$) and incurring a cost when approving harmful cases ($y{=}0$).

\subsection{Constraints}
We consider two classes of constraints, expressed via residuals $g(\theta)$ where feasibility means $g(\theta)\le 0$.

\textbf{Demographic Parity (DP).}
Let group-wise acceptance rates be
\begin{equation}
\mathrm{acc}_g(\theta)=\mathbb{P}(d=1 \mid a=g), \quad g\in\{0,1\}.
\end{equation}
The DP gap is $\Delta_{\mathrm{DP}}(\theta)=|\mathrm{acc}_0(\theta)-\mathrm{acc}_1(\theta)|$.
We enforce
\begin{equation}
g_{\mathrm{dp}}(\theta)=\Delta_{\mathrm{DP}}(\theta)-\varepsilon \le 0.
\end{equation}

\textbf{Service-rate constraint.}
To avoid degenerate always-reject or always-accept solutions, we optionally enforce an overall acceptance-rate window:
\begin{equation}
\mathrm{acc}(\theta)=\mathbb{P}(d=1) \in [\alpha_{\min}, \alpha_{\max}],
\end{equation}
encoded as two residuals $g_{\mathrm{srv,low}}(\theta)=\alpha_{\min}-\mathrm{acc}(\theta)$ and
$g_{\mathrm{srv,high}}(\theta)=\mathrm{acc}(\theta)-\alpha_{\max}$.

\subsection{Closed-loop dynamics and performance metrics}
Crucially, the environment is closed-loop: the distribution of $(s,y,a)$ at time $t$ depends on previous decisions, creating non-stationarity.
We evaluate:
\begin{itemize}
  \item \textbf{Tail reward / tail DP gap:} mean over the last $K_{\text{tail}}$ iterations ($K_{\text{tail}}{=}200$), then averaged across random seeds.
  \item \textbf{Cumulative violation:} sum over $t$ of the positive parts of residuals,
  \begin{equation}
  V_T = \sum_{t=1}^T \sum_j \max\{g_{j,t},0\}.
  \end{equation}
  This metric penalizes persistent or repeated constraint violations and is robust to stochastic fluctuations.
\end{itemize}
We also report bootstrap confidence intervals (CIs) across seeds for key differences (Table~\ref{tab:ci}).

% ===================== III. Optimistic Feasible Search =====================
\section{Optimistic Feasible Search (OFS)}\label{sec:ofs}
OFS targets low-dimensional policy classes where discretization is natural.
Let $\{\theta_i\}_{i=1}^{K_\theta}$ be a uniform grid on $[\theta_{\min},\theta_{\max}]$.
For each grid point, OFS maintains empirical means of reward and constraint residuals along with a confidence radius.

\subsection{Algorithm}
Let $\hat r_i(t)$ be the empirical mean reward for $\theta_i$ after it has been played $n_i(t)$ times.
Similarly, let $\hat g_{i,j}(t)$ denote the empirical mean of residual $j$.
We define a confidence radius
\begin{equation}
b_i(t)=\sqrt{\frac{c\,\log\big(\frac{(t+1)K_\theta}{\delta}\big)}{\max(1,n_i(t))}},
\end{equation}
with constant $c>0$ and confidence $\delta\in(0,1)$.

At round $t$, define the optimistic feasible set
\begin{equation}
\mathcal{F}_t = \Big\{ i: \hat g_{i,j}(t)+b_i(t)\le 0,\ \forall j \Big\}.
\end{equation}
OFS selects:
\begin{itemize}
\item If $\mathcal{F}_t \neq \emptyset$: choose $i_t \in \arg\max_{i\in\mathcal{F}_t} \big(\hat r_i(t)+b_i(t)\big)$.
\item Else: choose $i_t \in \arg\min_i \sum_j \max\big(\hat g_{i,j}(t)+b_i(t),0\big)$.
\end{itemize}
We also use a small $\epsilon_{\mathrm{exp}}$-greedy exploration probability to ensure coverage.

\begin{algorithm}[t]
\caption{Optimistic Feasible Search (OFS)}
\label{alg:ofs}
\begin{algorithmic}[1]
\STATE \textbf{Input:} grid $\{\theta_i\}_{i=1}^{K_\theta}$, confidence $\delta$, exploration $\epsilon_{\mathrm{exp}}$
\STATE Initialize $n_i \leftarrow 0$, $\hat r_i \leftarrow 0$, $\hat g_{i,j}\leftarrow 0$ for all $i,j$
\FOR{$t=1$ to $T$}
  \STATE Compute $b_i(t)=\sqrt{\frac{c\log(((t+1)K_\theta)/\delta)}{\max(1,n_i)}}$ for all $i$
  \STATE $\mathcal{F}_t \leftarrow \{ i : \hat g_{i,j}+b_i(t)\le 0\ \forall j \}$
  \STATE With prob.\ $\epsilon_{\mathrm{exp}}$, pick $i_t$ uniformly in $\{1,\dots,K_\theta\}$ (explore)
  \STATE Else if $\mathcal{F}_t\neq\emptyset$, pick $i_t\in\arg\max_{i\in\mathcal{F}_t} (\hat r_i+b_i(t))$
  \STATE Else pick $i_t\in\arg\min_i \sum_j \max(\hat g_{i,j}+b_i(t),0)$
  \STATE Play threshold $\theta_{i_t}$; observe reward $r_t$ and residuals $g_{t,j}$
  \STATE Update $n_{i_t}\leftarrow n_{i_t}+1$ and empirical means $\hat r_{i_t},\hat g_{i_t,j}$
\ENDFOR
\end{algorithmic}
\end{algorithm}

\subsection{Discussion and intuition}
OFS differs from standard primal--dual approaches in a crucial way:
instead of optimizing a Lagrangian with dual multipliers that can oscillate under non-stationarity, OFS explicitly screens feasibility using confidence bounds and focuses search on thresholds that are plausibly feasible.
When feasibility is currently uncertain, OFS naturally allocates samples to reduce uncertainty in regions that are near the boundary.

\subsection{Theoretical perspective (sketch)}
A full closed-loop analysis can be complex, but in our benchmarks the dynamics are designed to be stable and contractive.
This yields a useful perspective: each fixed threshold induces a well-defined steady-state distribution, and samples gathered when repeatedly choosing a threshold concentrate around its steady-state means.

\textbf{Assumption (contractive closed-loop).}
For each fixed threshold $\theta_i$, the environment state evolves via a contraction mapping with a unique stationary distribution, and observations have bounded noise (sub-Gaussian) around steady-state means.
This is satisfied by our synthetic benchmark by construction and approximately captured by the clipped, stable update rules in our semi-synthetic benchmark.

\textbf{Proposition (informal).}
Under the above assumption, with high probability $1-\delta$, OFS achieves (i) sublinear cumulative violation $V_T = \tilde O(\sqrt{T K_\theta})$ and (ii) near-optimal reward relative to the best feasible grid threshold, up to standard statistical and discretization errors.
The proof follows a standard optimism argument: confidence bounds hold uniformly over arms; OFS selects an arm whose optimistic feasibility implies true feasibility once sampled sufficiently; when infeasible, OFS minimizes optimistic violation, driving exploration toward feasibility.

We emphasize that our experiments validate these behaviors empirically and show that OFS is consistently near-oracle (Table~\ref{tab:oracle_gap}).

% ===================== IV. Baselines =====================
\section{Baselines}
We compare against two strong bandit baselines commonly used for constrained online optimization:

\textbf{Unconstrained bandit gradient descent (U).}
We apply a one-point zeroth-order gradient estimator \cite{flaxman2005online} to the loss $-\!r_t(\theta)$, ignoring constraints.

\textbf{Primal--Dual bandit gradient descent (PD).}
We optimize a Lagrangian with dual ascent on constraint residuals \cite{mahdavi2012online,achiam2017cpo}.
We include a small PD hyperparameter grid for the semi-synthetic tasks and report a tuning table to address baseline sensitivity concerns (Table~\ref{tab:pd_tuning}).

% ===================== V. Closed-loop Benchmarks =====================
\section{Closed-loop Benchmarks}\label{sec:benchmarks}

\subsection{Synthetic benchmark (MVE-S)}
We use a synthetic closed-loop environment where group-specific score distributions evolve with acceptance decisions.
At each round, a group $a\in\{0,1\}$ is sampled and the score is Gaussian with mean $\mu_a(t)$.
Labels are generated via a sigmoid model.
The decision threshold affects group acceptance rates, which then update the state $(\mu_0,\mu_1)$ through a stable contraction dynamics.
This benchmark isolates closed-loop feedback in a controlled setting while keeping the policy class (thresholds) simple.

\subsection{Semi-synthetic benchmarks (German Credit and COMPAS)}
To ground our closed-loop study in real tabular datasets, we construct semi-synthetic environments using German Credit and COMPAS \cite{dua2017uci,angwin2016machine}.
In German, we use sex as the sensitive attribute and define $y=1$ as good credit; in COMPAS, we use sex and define $y=1$ as non-recidivism.
We preprocess each dataset with one-hot encoding and standardization, then train a simple logistic score model to produce scalar logits $s$.
Closed-loop dynamics are induced by composition shift: for each group we maintain a mixture over a ``high-score'' pool and a ``low-score'' pool, and the mixture weight is updated as a stable function of group-wise acceptance rates.
Intuitively, higher acceptance can increase the fraction of ``high-score'' individuals observed in the future, capturing selection effects while remaining simple and reproducible.

% ===================== VI. Experiments =====================
\section{Experiments}\label{sec:experiments}

\subsection{Protocol and implementation details}
We run $10$ random seeds for each main experiment.
We report tail means over the last $K_{\text{tail}}{=}200$ iterations per seed, then report mean$\pm$std across seeds (Table~\ref{tab:main_results}).
Cumulative violation sums positive parts of constraint residuals over the full horizon (DP-only for MVE-S; DP + service constraints for German/COMPAS).
We also report bootstrap CIs (Table~\ref{tab:ci}).
Oracle tradeoff curves are computed by evaluating fixed thresholds over long horizons and selecting the best feasible point (Fig.~\ref{fig:oracle_curves}, Table~\ref{tab:oracle_gap}).

\subsection{Main results}
Table~\ref{tab:main_results} summarizes tail reward, tail DP gap, and cumulative violations on all three benchmarks.
OFS consistently achieves the highest reward while keeping DP gaps well below the required $\varepsilon$ and dramatically reducing cumulative violations.
On the semi-synthetic tasks, unconstrained learning incurs severe violations (due to both DP and service constraints), while PD improves utility but still exhibits substantially larger cumulative violation than OFS.

% ---------- Table I ----------
\begin{table*}[tb]
\centering
\small
\setlength{\tabcolsep}{4pt}
\caption{Main results. Metrics are computed as tail mean over the last $K_{\text{tail}}{=}200$ iterations per seed, then reported as mean$\pm$std over 10 seeds. Cumulative violation sums the positive parts of the constraints over the full horizon (DP only for MVE-S; DP + accept-rate constraints for German/COMPAS).}
\label{tab:main_results}
\begin{tabular}{llccc}
\toprule
Task & Algorithm & Reward $\uparrow$ & DP gap $\downarrow$ & Cum.\ viol.\ $\downarrow$ \\
\midrule
MVE-S ($\varepsilon=0.06$) & Unconstrained & 0.0657$\pm$0.0373 & 0.0953$\pm$0.0230 & 93.88$\pm$16.93 \\
 & PrimalDual & 0.0287$\pm$0.0155 & 0.0581$\pm$0.0141 & 33.14$\pm$9.96 \\
 & OFS-Grid & \textbf{0.3068}$\pm$0.0051 & \textbf{0.0125}$\pm$0.0014 & \textbf{7.98}$\pm$0.76 \\
\midrule
German ($\varepsilon=0.02$, $a\in[0.30,0.99]$) & Unconstrained & 0.1127$\pm$0.0306 & 0.0448$\pm$0.0078 & 622.10$\pm$18.63 \\
 & PrimalDual & 0.3929$\pm$0.0993 & 0.0309$\pm$0.0057 & 179.10$\pm$34.55 \\
 & OFS-Grid & \textbf{0.4546}$\pm$0.0064 & \textbf{0.0131}$\pm$0.0021 & \textbf{31.13}$\pm$1.54 \\
\midrule
COMPAS ($\varepsilon=0.03$, $a\in[0.30,0.99]$) & Unconstrained & 0.1377$\pm$0.0391 & 0.1399$\pm$0.0133 & 921.37$\pm$12.37 \\
 & PrimalDual & 0.6681$\pm$0.0225 & 0.0396$\pm$0.0046 & 161.49$\pm$12.92 \\
 & OFS-Grid & \textbf{0.7556}$\pm$0.0071 & \textbf{0.0101}$\pm$0.0023 & \textbf{26.53}$\pm$1.05 \\
\bottomrule
\end{tabular}
\end{table*}

\subsection{Learning dynamics on MVE-S}
\begin{figure*}[t]
\centering
\subfloat[Reward]{\includegraphics[width=0.32\textwidth]{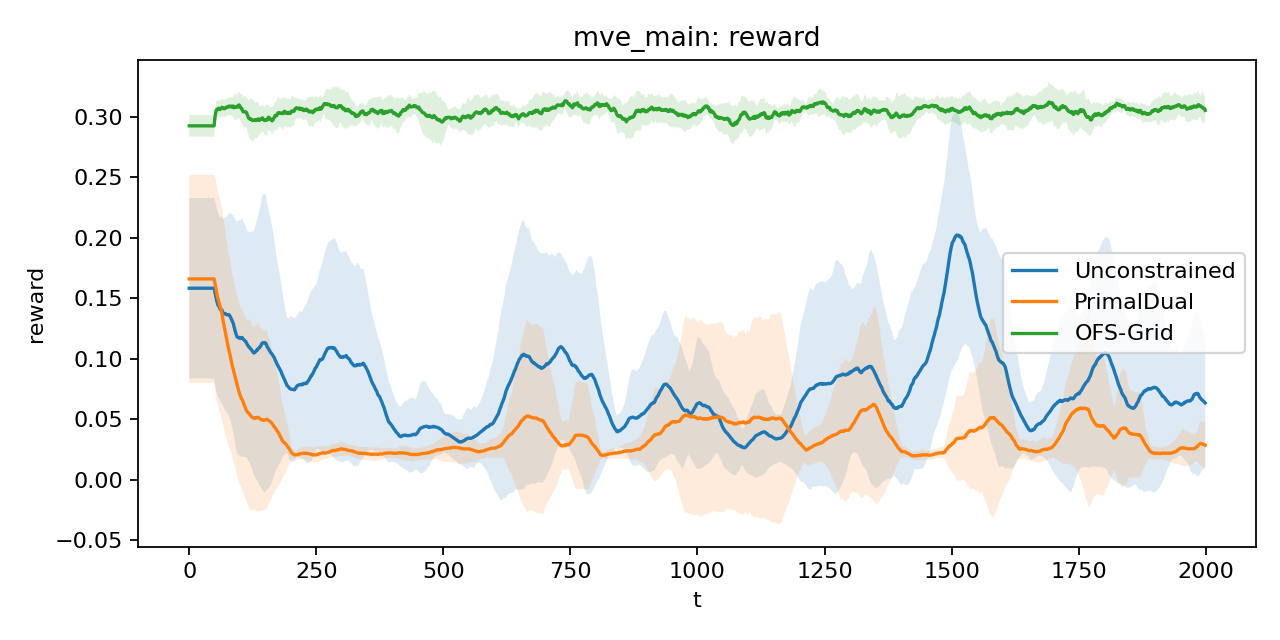}}
\hfill
\subfloat[DP gap]{\includegraphics[width=0.32\textwidth]{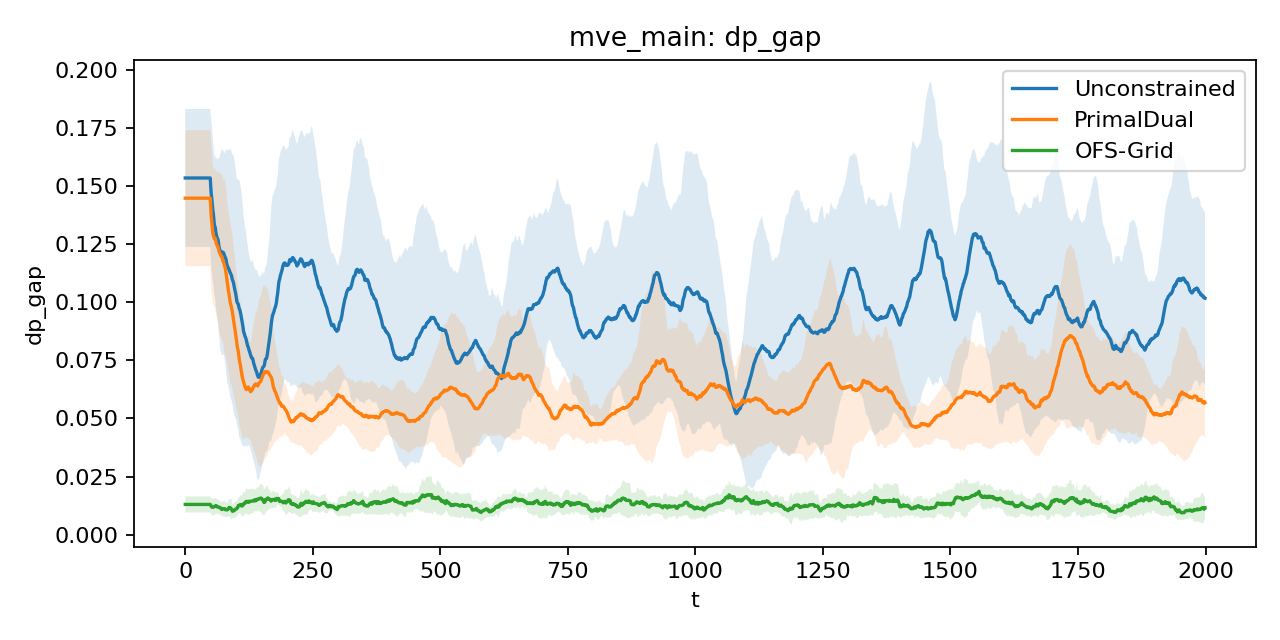}}
\hfill
\subfloat[Cumulative violation]{\includegraphics[width=0.32\textwidth]{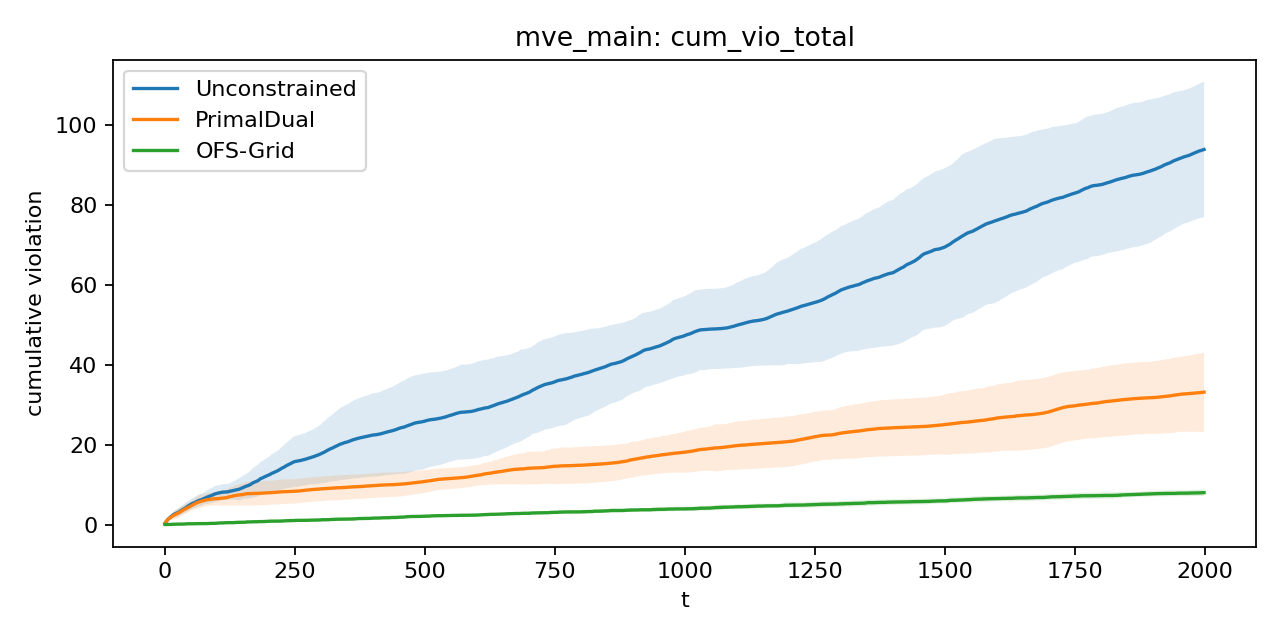}}
\caption{MVE-S closed-loop learning dynamics (mean$\pm$std over seeds). OFS achieves higher reward with dramatically smaller DP gap and cumulative violation.}
\label{fig:mve_curves}
\end{figure*}

Fig.~\ref{fig:mve_curves} shows representative learning curves on the synthetic benchmark.
OFS rapidly finds a feasible high-utility threshold: DP gap decreases and cumulative violations grow slowly, while reward remains high.
In contrast, the unconstrained baseline improves utility only when it allows large DP violations, and PD tends to trade reward for partial constraint control.

% ---------- Fig. 1 ----------

\subsection{Semi-synthetic learning dynamics}
\begin{figure*}[t]
\centering
\subfloat[German: reward]{\includegraphics[width=0.48\textwidth]{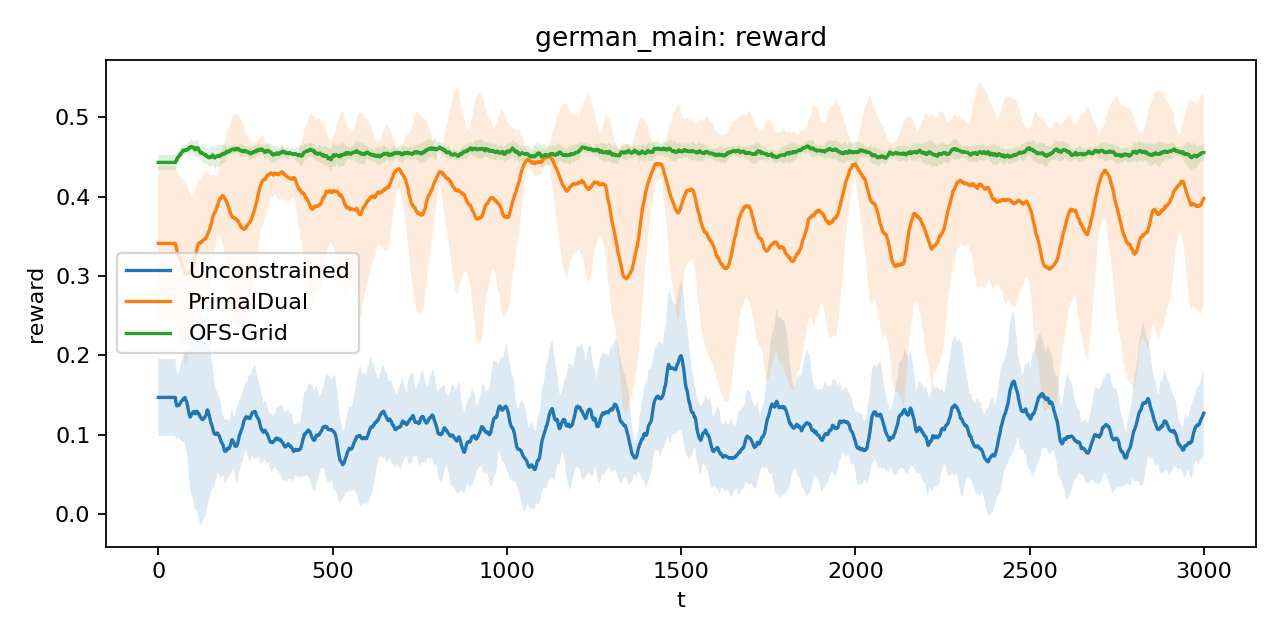}}
\hfill
\subfloat[German: DP gap]{\includegraphics[width=0.48\textwidth]{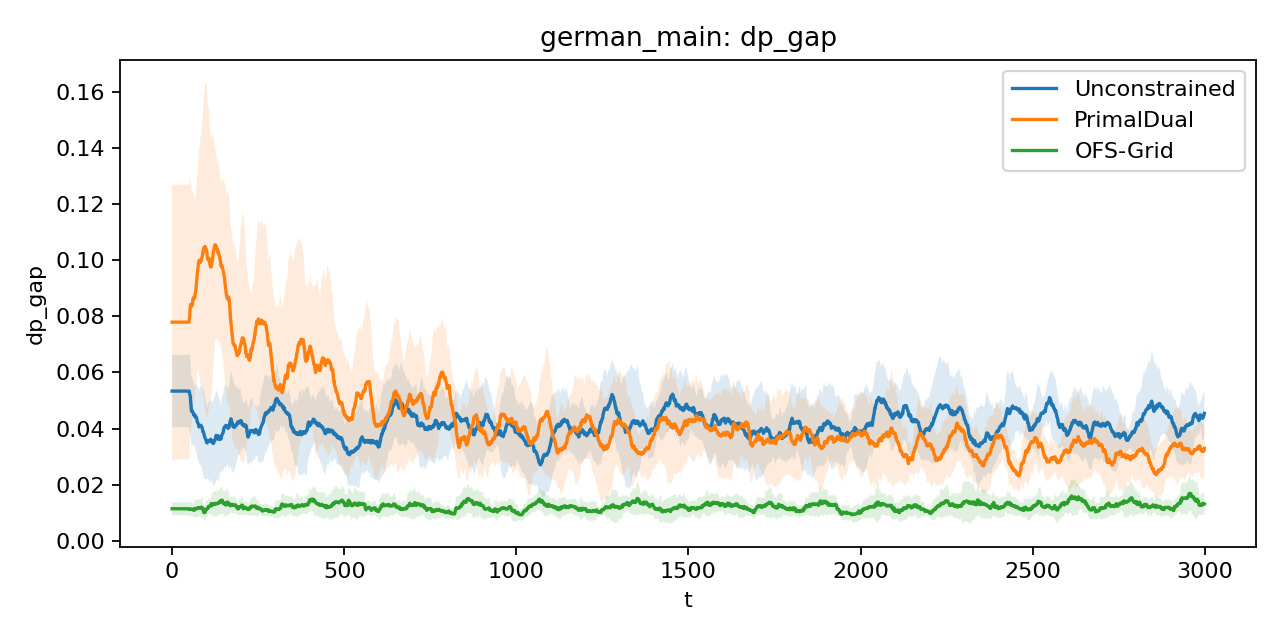}}\\
\subfloat[COMPAS: reward]{\includegraphics[width=0.48\textwidth]{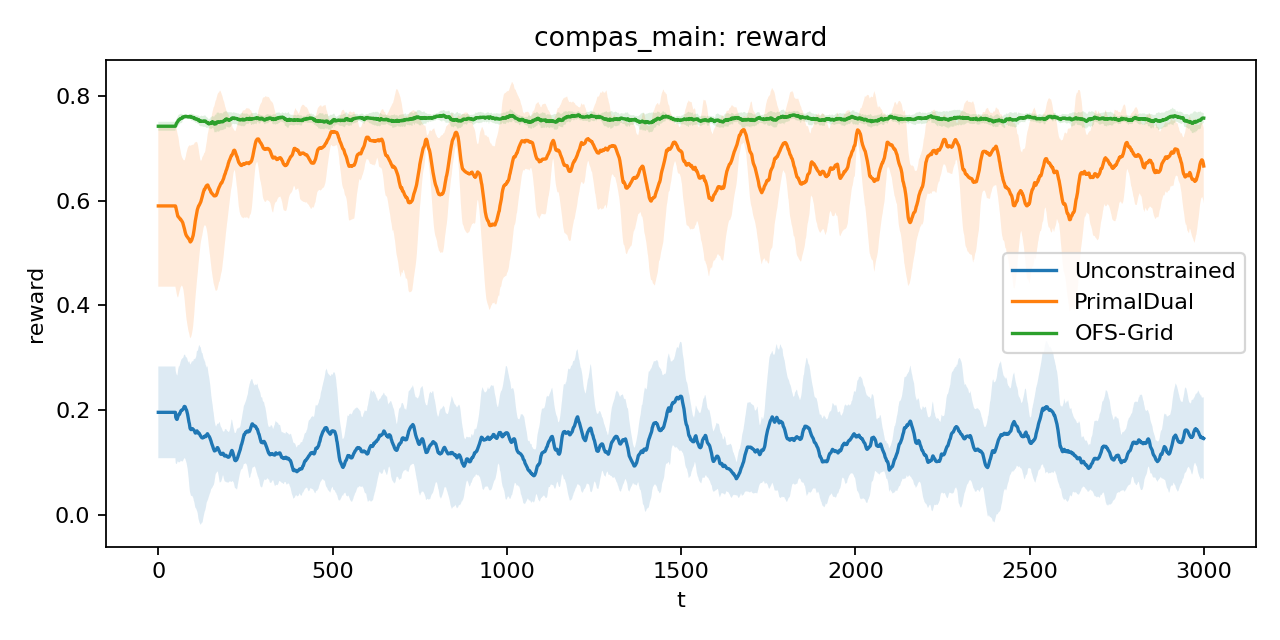}}
\hfill
\subfloat[COMPAS: DP gap]{\includegraphics[width=0.48\textwidth]{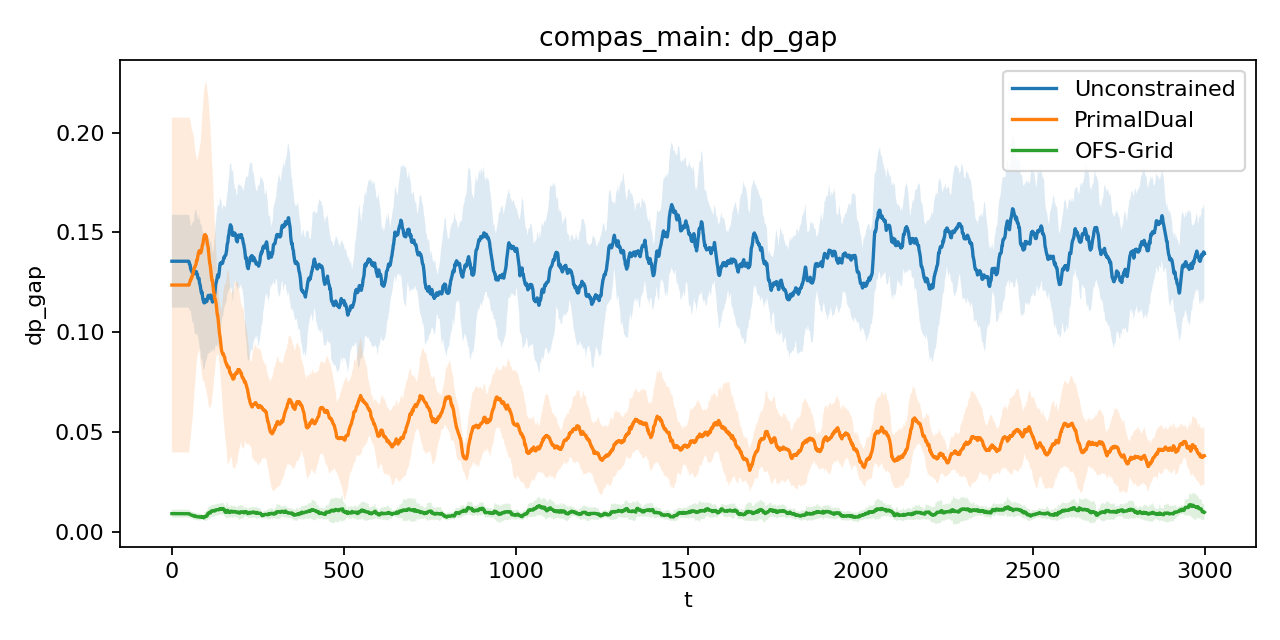}}
\caption{Semi-synthetic closed-loop learning curves on German Credit and COMPAS. OFS consistently reduces constraint violations while preserving high reward.}
\label{fig:semisyn_curves}
\end{figure*}

We next evaluate on German Credit and COMPAS semi-synthetic closed-loop environments.
Fig.~\ref{fig:semisyn_curves} shows that OFS keeps DP gaps small and service violations low, yielding far smaller cumulative violation while achieving high reward.
PD can achieve strong reward but exhibits substantially larger violation, especially under service constraints, which is precisely what our cumulative-violation metric captures.

% ---------- Fig. 2 ----------

\subsection{Near-oracle performance}
\begin{figure*}[t]
\centering
\subfloat[MVE-S]{\includegraphics[width=0.32\textwidth]{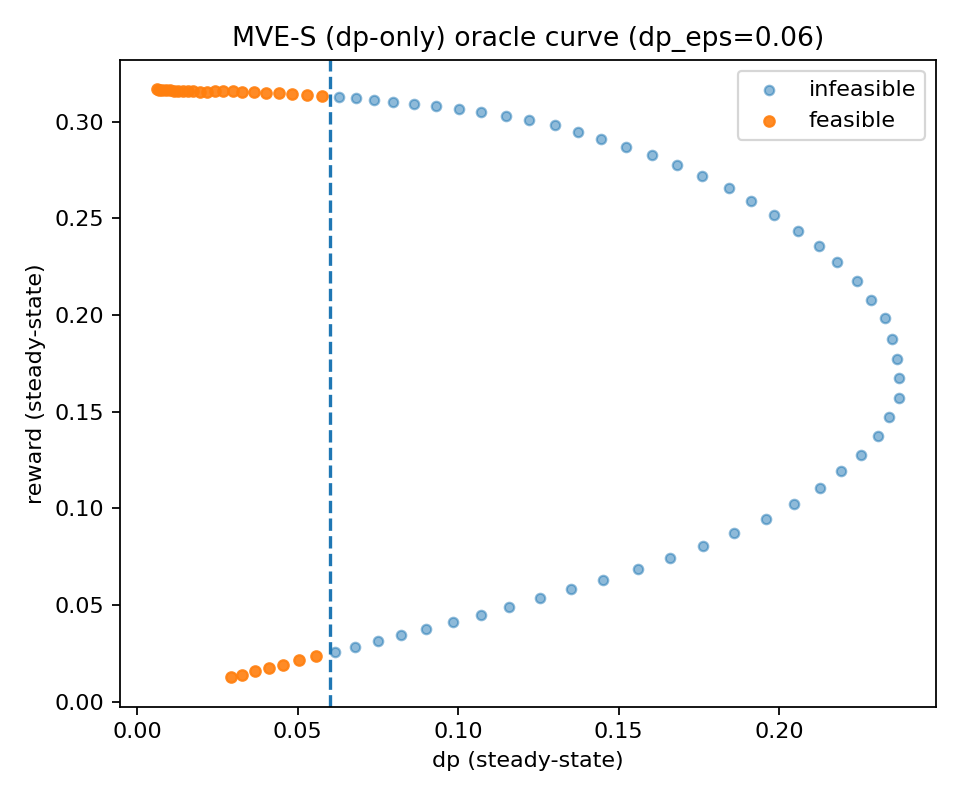}}
\hfill
\subfloat[German]{\includegraphics[width=0.32\textwidth]{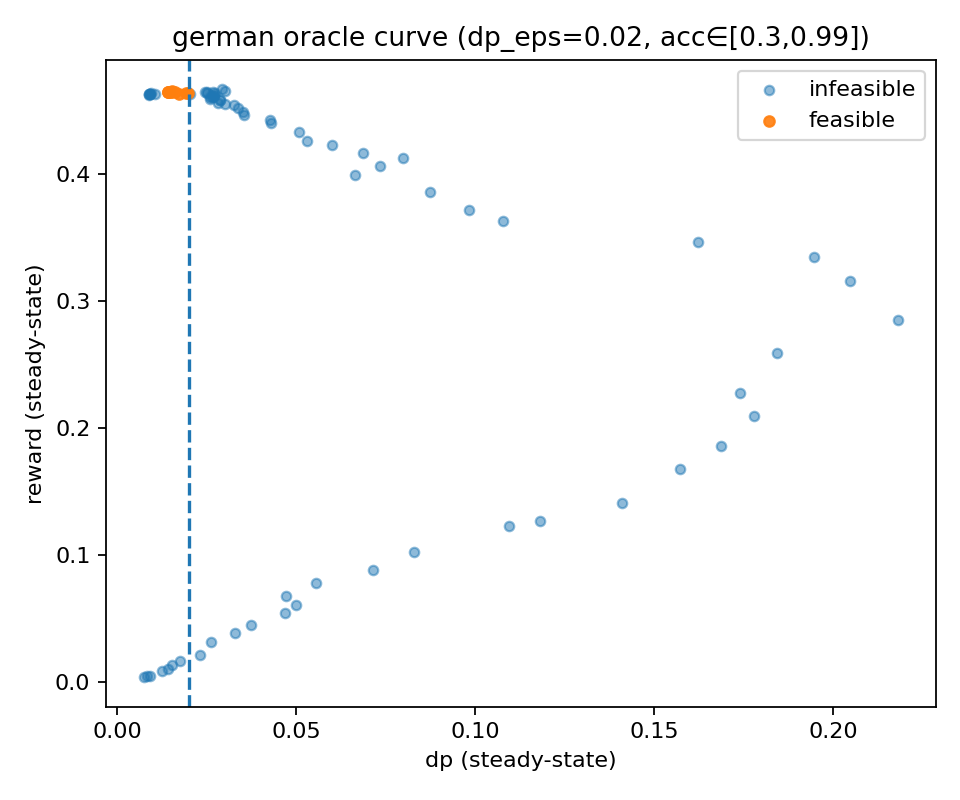}}
\hfill
\subfloat[COMPAS]{\includegraphics[width=0.32\textwidth]{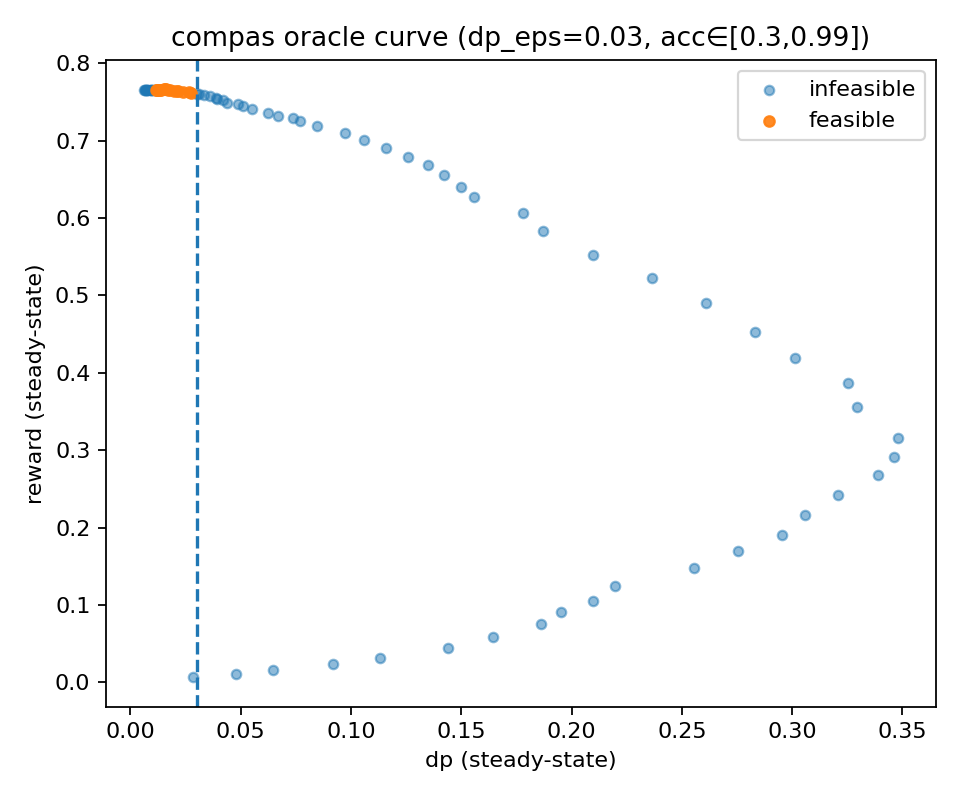}}
\caption{Oracle tradeoff curves (fixed threshold steady-state evaluation). Each curve is obtained by sweeping thresholds and estimating steady-state reward and DP gap; the oracle best feasible point is the feasible point with maximal reward. OFS operates online yet approaches oracle-quality feasible thresholds.}
\label{fig:oracle_curves}
\end{figure*}

To assess absolute optimality, we compute an oracle tradeoff curve by sweeping fixed thresholds and selecting the best feasible point.
Table~\ref{tab:oracle_gap} reports oracle best feasible reward and the steady-state suboptimality gaps.
Across all benchmarks, OFS is near-oracle: its reward gap to the best feasible fixed threshold is below $0.003$.
Fig.~\ref{fig:oracle_curves} visualizes the reward--DP tradeoff curves and highlights the feasible region.

% ---------- Table II ----------
\begin{table*}[tb]
\centering
\small
\setlength{\tabcolsep}{4pt}
\caption{Oracle best feasible performance (steady-state) and suboptimality gaps. Oracle is obtained by sweeping fixed thresholds $\theta$ and selecting the best feasible point; gaps are oracle reward minus each algorithm's steady-state reward (smaller is better).}
\label{tab:oracle_gap}
\begin{tabular}{lcccccc}
\toprule
Task & Oracle reward & Oracle DP & Oracle acc & Gap(U) & Gap(PD) & Gap(OFS) \\
\midrule
MVE-S & 0.3168 & 0.0062 & 0.9964 & 0.2571 & 0.2911 & \textbf{0.0026} \\
German & 0.4656 & 0.0153 & 0.9877 & 0.3303 & 0.0382 & \textbf{0.0028} \\
COMPAS & 0.7665 & 0.0157 & 0.9849 & 0.5950 & 0.0050 & \textbf{0.0012} \\
\bottomrule
\end{tabular}
\end{table*}

% ---------- Fig. 3 ----------

\subsection{Statistical significance via bootstrap CIs}
We compute bootstrap confidence intervals across seeds for key performance differences (e.g., OFS reward improvement over PD; PD vs.\ OFS in DP gap and violation).
Table~\ref{tab:ci} shows that OFS improvements are statistically robust across benchmarks.

% ---------- Table III ----------
\begin{table}[t]
\centering
\small
\setlength{\tabcolsep}{3pt}
\renewcommand{\arraystretch}{1.1}
\caption{Bootstrap confidence intervals (95\%) for key differences, computed over seeds using paired bootstrap.}
\label{tab:ci}
\begin{tabular}{lccc}
\toprule
Task & \shortstack{$\Delta$Reward\\(OFS--PD)} & \shortstack{$\Delta$DP\\(PD--OFS)} & \shortstack{$\Delta$Viol.\\(PD--OFS)} \\
\midrule
MVE-S & \shortstack{0.2781\\ {[0.2683, 0.2858]}} & \shortstack{0.0457\\ {[0.0375, 0.0544]}} & \shortstack{25.16\\ {[20.05, 32.29]}} \\
German & \shortstack{0.0617\\ {[0.0051, 0.1261]}} & \shortstack{0.0178\\ {[0.0149, 0.0211]}} & \shortstack{147.98\\ {[127.88, 170.01]}} \\
COMPAS & \shortstack{0.0875\\ {[0.0714, 0.1027]}} & \shortstack{0.0295\\ {[0.0263, 0.0326]}} & \shortstack{134.96\\ {[127.30, 142.90]}} \\
\bottomrule
\end{tabular}
\end{table}

\subsection{Baseline tuning: Primal--Dual hyperparameters}
A common critique for primal--dual baselines is sensitivity to step sizes and dual updates.
We therefore include a small PD tuning grid and report results (Table~\ref{tab:pd_tuning}).
Even after tuning, OFS maintains superior reward--violation tradeoffs on the semi-synthetic tasks.
We will release the full tuning grid and experiment logs upon publication.

% ---------- Table IV ----------
\begin{table}[t]
\centering
\small
\setlength{\tabcolsep}{3pt}
\renewcommand{\arraystretch}{1.05}
\caption{PrimalDual (PD) hyperparameter tuning for semi-synthetic tasks (small grid).}
\label{tab:pd_tuning}
\begin{tabular}{lc>{\raggedright\arraybackslash}p{0.42\columnwidth}}
\toprule
Dataset & Selected $(\eta,\mu)$ & Selection rule \\
\midrule
German & (0.03, 0.10) & infeasible-min-violation \\
COMPAS & (0.03, 0.40) & infeasible-min-violation \\
\bottomrule
\end{tabular}
\end{table}

\subsection{Robustness: stress sweeps and strict infeasibility}
We evaluate robustness by sweeping key parameters (cost, fairness tolerance, feedback strength) and computing win-rates for OFS under feasibility-aware selection rules.
Table~\ref{tab:stress_winrate} reports stress win-rates.
For each stress configuration, we declare the winner as the feasible algorithm with the highest tail reward; if no algorithm is feasible, the winner is the one with the smallest cumulative violation.
We also include a strict infeasible configuration where constraints are intentionally conflicting; in this regime, no method can be perfectly feasible, so the goal is to minimize cumulative violation while maintaining utility.
Table~\ref{tab:strict_infeasible} summarizes strict infeasible results.

% ---------- Table V and VI ----------
\begin{table}[t]
\centering
\small
\caption{Stress sweep win rate (percentage of stress settings where the method is the winner). Values are taken from stress\_winrate\_*.json; missing entries are reported as 0.0.}
\label{tab:stress_winrate}
\begin{tabular}{lcc}
\toprule
Task & OFS-Grid & PrimalDual \\
\midrule
MVE-S & 88.9\% & 11.1\% \\
German & 100.0\% & 0.0\% \\
COMPAS & 100.0\% & 0.0\% \\
\bottomrule
\end{tabular}
\end{table}

\begin{table}[t]
\centering
\small
\setlength{\tabcolsep}{3pt}
\renewcommand{\arraystretch}{1.05}
\caption{Strict infeasible regime with tighter constraints ($\varepsilon=0.01$, $\mathrm{acc}\in[0.35,0.85]$) for semi-synthetic tasks. Metrics are reported using the same protocol as the main results.}
\label{tab:strict_infeasible}
\resizebox{\columnwidth}{!}{%
\begin{tabular}{llccc}
\toprule
Setting & Algorithm & Reward $\uparrow$ & DP gap $\downarrow$ & Cum.\ viol.\ $\downarrow$ \\
\midrule
German (strict) & Unconstrained & 0.1127$\pm$0.0306 & 0.0448$\pm$0.0078 & 797.41$\pm$17.95 \\
 & PrimalDual & 0.4115$\pm$0.0138 & 0.0469$\pm$0.0064 & \textbf{383.10}$\pm$14.02 \\
 & OFS-Grid & \textbf{0.4546}$\pm$0.0064 & \textbf{0.0131}$\pm$0.0021 & 445.69$\pm$0.92 \\
\midrule
COMPAS (strict) & Unconstrained & 0.1377$\pm$0.0391 & 0.1399$\pm$0.0133 & 1130.38$\pm$11.32 \\
 & PrimalDual & 0.6519$\pm$0.0251 & 0.0408$\pm$0.0043 & 476.69$\pm$3.81 \\
 & OFS-Grid & \textbf{0.7556}$\pm$0.0071 & \textbf{0.0101}$\pm$0.0023 & \textbf{440.62}$\pm$1.09 \\
\bottomrule
\end{tabular}
}
\end{table}

% ===================== VII. Discussion and Limitations =====================
\section{Discussion and Limitations}
\textbf{Why OFS works well here.}
Closed-loop fairness problems often suffer from instability and oscillations when constraints are enforced indirectly.
OFS explicitly searches for thresholds that appear feasible under uncertainty, which is particularly effective for low-dimensional, interpretable policy classes (thresholds) where discretization provides a transparent search space.
In our experiments, OFS quickly identifies the feasible region and then refines reward within it, matching near-oracle performance.

\textbf{Limitations.}
First, our current focus is on one-dimensional thresholds; extending optimism-based feasible screening to higher-dimensional policy classes may require structured discretization or surrogate modeling.
Second, we focus on DP (and a service-rate constraint); other fairness notions (e.g., equalized odds \cite{hardt2016equality}) may require different constraint estimators and could change tradeoffs.
Third, our semi-synthetic benchmark captures composition shift via a simple feedback model; real-world feedback mechanisms can be more complex and may involve strategic behavior \cite{perdomo2020performative}.

% ===================== VIII. Conclusion =====================
\section{Conclusion}
We presented OFS, an optimism-based method for closed-loop threshold learning under constraints.
Across a synthetic benchmark and two semi-synthetic benchmarks derived from German Credit and COMPAS, OFS achieves higher reward and lower cumulative violations than strong baselines while remaining near-oracle.
Our pipeline is fully reproducible and uses double-blind friendly relative outputs, making it straightforward to validate and extend these results.

\bibliographystyle{IEEEtran}
\bibliography{refs}

@inproceedings{dwork2012fairness,
  title={Fairness through awareness},
  author={Dwork, Cynthia and Hardt, Moritz and Pitassi, Toniann and Reingold, Omer and Zemel, Richard},
  booktitle={Proceedings of the 3rd Innovations in Theoretical Computer Science Conference (ITCS)},
  address={Cambridge, MA, USA},
  month=jan,
  year={2012},
  pages={214--226}
}

@inproceedings{feldman2015certifying,
  title={Certifying and removing disparate impact},
  author={Feldman, Michael and Friedler, Sorelle A. and Moeller, John and Scheidegger, Carlos and Venkatasubramanian, Suresh},
  booktitle={Proceedings of the 21st ACM SIGKDD International Conference on Knowledge Discovery and Data Mining (KDD)},
  address={Sydney, NSW, Australia},
  month=aug,
  year={2015},
  pages={259--268}
}

@inproceedings{agarwal2018reductions,
  title={A reductions approach to fair classification},
  author={Agarwal, Alekh and Beygelzimer, Alina and Dud\'{\i}k, Miroslav and Langford, John and Wallach, Hanna},
  booktitle={Proceedings of the 35th International Conference on Machine Learning (ICML)},
  address={Stockholm, Sweden},
  month=jul,
  year={2018},
  pages={60--69}
}

@inproceedings{hardt2016equality,
  title={Equality of opportunity in supervised learning},
  author={Hardt, Moritz and Price, Eric and Srebro, Nati},
  booktitle={Advances in Neural Information Processing Systems 30},
  address={Barcelona, Spain},
  month=dec,
  year={2016},
  pages={3323--3331}
}

@inproceedings{kleinberg2017inherent,
  title={Inherent trade-offs in the fair determination of risk scores},
  author={Kleinberg, Jon and Mullainathan, Sendhil and Raghavan, Manish},
  booktitle={Proceedings of the 8th Innovations in Theoretical Computer Science Conference (ITCS)},
  series={LIPIcs},
  volume={67},
  address={Berkeley, CA, USA},
  month=jan,
  year={2017},
  pages={43:1--43:23}
}

@article{auer2002ucb,
  title={Finite-time analysis of the multi-armed bandit problem},
  author={Auer, Peter and Cesa-Bianchi, Nicolo and Fischer, Paul},
  journal={Machine Learning},
  volume={47},
  number={2--3},
  pages={235--256},
  month=may,
  year={2002}
}

@inproceedings{zinkevich2003online,
  title={Online convex programming and generalized infinitesimal gradient ascent},
  author={Zinkevich, Martin},
  booktitle={Proceedings of the 20th International Conference on Machine Learning (ICML)},
  address={Washington, DC, USA},
  month=aug,
  year={2003},
  pages={928--936}
}

@article{mahdavi2012online,
  title={Trading regret for efficiency: online convex optimization with long-term constraints},
  author={Mahdavi, Mehrdad and Jin, Rong and Yang, Tianbao},
  journal={Journal of Machine Learning Research},
  volume={13},
  pages={2503--2528},
  year={2012}
}

@inproceedings{achiam2017cpo,
  title={Constrained policy optimization},
  author={Achiam, Joshua and Held, David and Tamar, Aviv and Abbeel, Pieter},
  booktitle={Proceedings of the 34th International Conference on Machine Learning (ICML)},
  address={Sydney, NSW, Australia},
  month=aug,
  year={2017},
  volume={70},
  pages={22--31}
}

@inproceedings{flaxman2005online,
  title={Online convex optimization in the bandit setting: gradient descent without a gradient},
  author={Flaxman, Abraham D. and Kalai, Adam Tauman and McMahan, H. Brendan},
  booktitle={Proceedings of the 16th Annual ACM-SIAM Symposium on Discrete Algorithms (SODA)},
  address={Vancouver, BC, Canada},
  month=jan,
  year={2005},
  pages={385--394}
}

@article{spall1992spsa,
  title={Multivariate stochastic approximation using a simultaneous perturbation gradient approximation},
  author={Spall, James C.},
  journal={IEEE Transactions on Automatic Control},
  volume={37},
  number={3},
  pages={332--341},
  month=mar,
  year={1992}
}

@article{nesterov2017random,
  title={Random gradient-free minimization of convex functions},
  author={Nesterov, Yurii and Spokoiny, Vladimir},
  journal={Foundations of Computational Mathematics},
  volume={17},
  number={2},
  pages={527--566},
  year={2017}
}

@inproceedings{perdomo2020performative,
  title={Performative prediction},
  author={Perdomo, Juan and \v{Z}rni\'c, Tijana and Mendler-D\"unner, Celestine and Hardt, Moritz},
  booktitle={Proceedings of the 37th International Conference on Machine Learning (ICML)},
  address={Virtual (originally Vienna, Austria)},
  month=jul,
  year={2020},
  volume={119},
  pages={7599--7609}
}

@misc{dua2017uci,
  title={UCI Machine Learning Repository},
  author={Dua, Dheeru and Graff, Casey},
  howpublished={University of California, Irvine},
  year={2017},
  note={Accessed Dec. 24, 2025}
}

@misc{angwin2016machine,
  title={Machine bias: There's software used across the country to predict future criminals. And it's biased against blacks},
  author={Angwin, Julia and Larson, Jeff and Mattu, Surya and Kirchner, Lauren},
  howpublished={ProPublica},
  month=may,
  year={2016},
  note={Accessed Dec. 24, 2025}
}
\end{document}